\pdfoutput=1

\documentclass[11pt]{article}
\usepackage{amsmath}
\usepackage[final]{acl}

\usepackage{times}
\usepackage{latexsym}
\usepackage{booktabs}
\usepackage{multirow}
\usepackage[T1]{fontenc}

\usepackage[utf8]{inputenc}

\usepackage{microtype}

\usepackage{inconsolata}

\usepackage{graphicx}

\title{Semantic Pivots Enable Cross-Lingual Transfer in Large Language Models}

\author{
Kaiyu He$^{1,3}$, 
Tong Zhou$^1$, 
Yubo Chen$^{1,2}$, 
Delai Qiu$^4$,
Shengping Liu$^4$,
Kang Liu$^{1,2,5}$,
Jun Zhao$^{1,2}$\\
$^1$The Key Laboratory of Cognition and Decision Intelligence for Complex Systems \\
Institute of Automation, Chinese Academy of Sciences\\
$^2$School of Artificial Intelligence, University of Chinese Academy of Sciences\\
$^3$University of Science and Technology Beijing\\
$^4$Unisound Al Technology Co,Ltd    $\quad^5$Shanghai Artificial Intelligence Laboratory\\
kaiyu\_he398@163.com,
tong.zhou@ia.ac.cn\\
\{yubo.chen, kliu, jzhao\}@nlpr.ia.ac.cn, \{qiudelai, liushengping\}@unisound.com
}

\begin{document}
\maketitle
\begin{abstract}

Large language models (LLMs) demonstrate remarkable ability in cross-lingual tasks. Understanding how LLMs acquire this ability is crucial for their interpretability. To quantify the cross-lingual ability of LLMs accurately, we propose a Word-Level Cross-Lingual Translation Task. To find how LLMs learn cross-lingual ability, we trace the outputs of LLMs' intermediate layers in the word translation task. We identify and distinguish two distinct behaviors in the forward pass of LLMs: co-occurrence behavior and semantic pivot behavior. We attribute LLMs' two distinct behaviors to the co-occurrence frequency of words and find the semantic pivot from the pre-training dataset. Finally, to apply our findings to improve the cross-lingual ability of LLMs, we reconstruct a semantic pivot-aware pre-training dataset using documents with a high proportion of semantic pivots. Our experiments validate the effectiveness of our approach in enhancing cross-lingual ability. Our research contributes insights into the interpretability of LLMs and offers a method for improving LLMs' cross-lingual ability.

\end{abstract}

\section{Introduction}

Multilingual large language models (MLLMs) demonstrate excellent cross-lingual ability \citep{le2023bloom, llama3modelcard, dang2024ayaexpansecombiningresearch} in various cross-lingual tasks, even in some low-resource languages \citep{lai2023chatgpt}. However, several critical questions remain unresolved: how to quantify the cross-lingual abilities of LLMs, how the cross-lingual abilities are established, and how to enhance the model's cross-lingual abilities. 

Quantifying the cross-lingual ability of LLMs is crucial to understanding how they acquire cross-lingual ability. Prior evaluations of LLMs have primarily focused on multilingual capabilities \citep{lewis2019mlqa}. Existing studies evaluating the cross-lingual abilities mainly focus on machine translation tasks \citep{flores101}. The discontinuity of cross-lingual ability scores measured in sentence-level translation tasks hinders the analysis of the model's cross-lingual ability, especially for early checkpoints with limited or poor linguistic capabilities. To address this shortcoming, our work introduces a word-level translation task and proposes a continuous metric from LLMs' output logits to accurately quantify their cross-lingual ability. 

The interpretability of LLMs is crucial for explaining how LLMs learn cross-lingual ability. \citet{chen2024rise} describes the entire process of LLMs acquiring new language capabilities. \citet{wendler2024llamas} discover that Llama-2 models internally use English as a pivot language. \citet{zhong2024beyond} further demonstrates that the LLMs' pivotal language is correlated with the pre-training corpus. However, these works do not attribute the LLMs' cross-lingual ability to the details of the pre-training corpus. We aim to explore the intrinsic relationship between LLMs' cross-lingual ability and the content of the pre-training corpus. 

It remains a challenge to propose a method that can effectively improve cross-lingual ability. Previous studies utilize extensive parallel corpora to improve LLMs' cross-lingual ability \citep{yang2023bigtranslate}. However, using excessive parallel data can have adverse effects, potentially due to catastrophic forgetting \citep{xu2023paradigm}. Based on our findings, we aim to effectively improve cross-lingual ability by reconstructing a pre-training dataset.

In this paper, to quantify the cross-lingual ability of LLMs accurately, we propose a \textbf{Cross-Lingual Word Translation Dataset (CLWTD)} to evaluate LLMs' cross-lingual ability, including four languages and a corresponding metric. By focusing on word-level translation tasks, we simplify the task of analyzing LLMs' internal states and pre-training corpus search. We assess the cross-lingual ability of LLMs with different sizes and series and acquire corresponding cross-lingual ability score matrices.
To explain how models acquire cross-lingual ability, we analyze the output of LLMs' intermediate layers using logit lens and discover LLMs' \textbf{two distinct behaviors}: co-occurrence behavior and semantic pivot behavior. Different from \citet{wendler2024llamas}, we focus on the relation between intermediate layer outputs, the input, and the final output. With the help of the tools WIMBD \citep{elazar2023s} and Infini-gram \citep{Liu2024InfiniGram}, we obtain the frequency of strings' co-occurrence in the pre-training dataset and the corresponding documents. We attribute LLMs' two behaviors to the co-occurrence frequency of strings. To find the semantic pivots from the pre-training dataset when LLM exhibits the semantic pivot behavior, we provide a method to select the tokens with a high relative co-occurrence proportion with both the source word and the target word. 
Finally, to apply our findings to enhancing LLMs' cross-lingual ability, we construct a \textbf{dataset with a high proportion of semantic pivots}. We build the adjacency matrix to obtain a set of semantic pivots and select our pre-training data by ranking documents based on the proportion of the tokens that belong to a set of semantic pivots. We train the OLMo-1B model in our dataset and compare it to two baseline models, demonstrating the effectiveness of our approach in improving cross-lingual ability.

In summary, our key contributions are as follows: 
\begin{itemize}
\item We propose a Cross-Lingual Word Translation Dataset (CLWTD) to evaluate LLMs' cross-lingual ability. Furthermore, we compare our metrics with chrF++, and the results demonstrate that our method can precisely reflect the cross-lingual ability of models.
\item We identify and distinguish LLMs' two behaviors. We attribute LLMs’ two distinct behaviors to the co-occurrence frequency of words and metric AUC score. We find the semantic pivot from the pre-training dataset and discover that its probability exhibits a trend of initially increasing and then decreasing.
\item We construct a semantic pivot-aware pre-training dataset and effectively improve the model's cross-lingual ability. The model trained on our dataset achieves a 0.013 improvement over the original published checkpoint model and a 0.005 improvement over the multilingual dataset baseline. 
\end{itemize}

\section{Related Work}

\subsection{Multilingual Evaluation}
Multilingual large language models (MLLMs) 
are trained on datasets comprising multiple languages, including mBERT \citep{devlin2018bert}, mBART \citep{liu2020multilingual}, XLM-R \citep{DBLP:journals/corr/abs-1911-02116}, mT5 \citep{xue-etal-2021-mt5}, XGLM \citep{lin-etal-2022-shot}, Aya Expanse \citep{dang2024ayaexpansecombiningresearch}, BLOOM \citep{le2023bloom}. These MLLMs are evaluated by various tasks, including classification task \citep{conneau2018xnli, ponti2020xcopa, lin-etal-2022-shot}, question answering \citep{artetxe-etal-2020-cross}, and natural language generation \citep{hasan-etal-2021-xl}. MEGA is the first multilingual benchmark for evaluating generative AI \citep{ahuja-etal-2023-mega}. \citet{lai-etal-2023-chatgpt} provides a multilingual holistic evaluation in 7 diverse NLP tasks. \citet{qi2023cross} measures the cross-lingual consistency of the models. For generative tasks, existing works primarily rely on string comparison as the metric for evaluation, like BLUE \citep{papineni-etal-2002-bleu} and CHRF \citep{popovic2015chrf}.

To improve MLLM's performance in this task, \citet{wei2023polylm, sanh2021multitask} fine-tune MLLMs using samples of various NLP tasks. They enhance the performance of MLLMs on unseen tasks. \citet{shi2024continual,mu2024revealing}
do continual pre-training to enhance MLLMs' ability to perform specific tasks. \citet{huang20241+}
transforming the input language of a model effectively improves MLLMs' performance. \citet{zhang2023plug,upadhayay2023taco} train MLLMs to process non-English instructions by initially interpreting them in English before generating a response in the target language.

\subsection{Interpretability of Multilingual Large Language Models}

Understanding how MLLMs handle multilingual capabilities and process inputs from different languages plays a crucial role in the interpretability of MLLMs \citep{zhao2024large}. \citet{blevins-etal-2022-analyzing,chen2024rise} explain the patterns in the dynamic training process of MLLMs. Some works explore models through neuron-level interpretation. \citet{zhang2024unveiling} finds a core region of multilingual capability. \citet{tang2024language,liu2024unraveling} identify language-agnostic neurons and language-specific neurons. \citet{bhattacharya-bojar-2023-unveiling} investigating component-based interpretation discovered distinct patterns of multilingual processing of the model's feed-forward network in the sub-layers. 
Parametric Probing Interpretation is widely used in the interpretability of multilingual models, such as probing classifiers \citep{starace-etal-2023-probing}, cross-lingual fine-tuning \citep{vulic-etal-2023-probing}, and linear probes \citep{de-varda-marelli-2024-emergence}. 
Several studies analyze the language behavior of MLLMs during the intermediate stages of static reasoning \citep{bhattacharya-bojar-2023-unveiling}. \citet{wendler2024llamas} discover that the language transformation occurs in the intermediate layers, when Llama-2 does the non-English tasks. They apply a tool, \textit{logit lens} \citep{logit-lens}, used to decode the model's output of the intermediate layer. By using the logit lens, the outputs of an intermediate layer forward pass the model's head directly, which is applied in the final layer of LLM, and get the logit of the outputs. The probability distribution of the intermediate layers can be obtained using softmax on the logit.


However, prior work does not focus on the role of pre-training datasets in establishing LLMs' cross-lingual capabilities.

\section{Cross-Lingual Ability}
Existing approaches to measuring cross-lingual ability including machine translation \citep{flores101,zhang-etal-2020-improving,ponti2020xcopa}, question answering \citep{lewis2019mlqa}, commonsense reasoning \citep{DBLP:journals/corr/abs-2112-10668}, etc. However, these sentence-level evaluation methods rely on discrete decoding and string matching, making it difficult to discern subtle differences in cross-lingual ability between models. In contrast, the more atomic and continuous word-level logits can not only faithfully reflect the cross-lingual abilities of models but also discern tiny differences between these abilities. Thus, we propose the Cross-Lingual Word Translation Dataset (CLWTD) and a corresponding metric to evaluate LLMs' cross-lingual ability.

\subsection{Word-Level Cross-Lingual Translation Task}
\label{3.1}
To evaluate the cross-lingual word-level capabilities of LLMs, we propose a Cross-Lingual Word Translation Dataset (CLWTD), which includes 2,000 parallel words and four languages. It offers strong scalability and enables more precise measurement of cross-lingual ability by mapping it into a continuous scale. This allows for the accurate measurement of even weakly performing LLMs.

We evaluate LLMs' cross-lingual ability on a word translation task using the CLWTD, the details shown in Appendix \ref{a}. Given a source language word, the task requires the MLLM to generate its translation in the target language. We calculate the loss when LLMs generate the target language word $y$ with $N$ tokens and transfer it to the possibility $P(y)$. 

$$loss(y)=-\frac{1}{N}\sum_{i = 1}^{N}\log(P(y_{i}))$$
$$P(y) = e^{-loss(y)}$$
$P(y_i)$ denotes the probability when LLMs generate the i-th token $y_i$ of the target language word.

We observe a continuous decrease in loss as the model generates a sequence of tokens. To mitigate potential biases introduced by long target word tokens, we introduce a \textbf{continuous output probability calibration} method. We design nine distractors $w' \in D'$ based on the source language word $w$ to evaluate and adjust the bias. Following the method for calculating the probability using the source language word $P(w)$, we calculate the probability of each distractor being translated into the target language word $P(w')$ by only replacing the source language words with distractors, quantifying the bias introduced by sequentially generated tokens. The cross-lingual ability score is measured by subtracting the average probability of the distractors from the probability of the source language word. Cross-lingual ability is defined as the mean value of these scores.
 
$$Score(w) = P(w) - \frac{\sum_{w'\in D'} P(w')}{|D'|}$$

In the zero-shot setting, LLMs struggle to consistently generate outputs in the desired format and exhibit relatively limited performance. By adopting a few-shot learning strategy, LLMs can imitate the format of shots, leading to more consistent output generation. In order to enable the model to adapt to the word translation task, when calculating the loss of the word translation task each time, we designed a five-shot prompt template shown in Appendix \ref{b} and randomly selected five words with their correct translations from the dataset as examples. To reduce the fluctuation introduced by examples, we measure five times for each word and calculate the average score.

We present a scalable approach for constructing our dataset. Our dataset encompasses four languages: English, Chinese, French, and Japanese. English and Chinese are chosen due to their dominance in the pre-training corpora of LLM. Additionally, we include French and Japanese due to their structural similarities to Chinese and English, respectively. Firstly, we select English words from the Oxford Dictionary and then translate them into the other four languages using Microsoft's translation tool \footnote{\url{https://www.bing.com/translator}}. Then, we utilize DeepSeek \footnote{\url{https://www.deepseek.com/}} to generate distractors that exhibit a degree of similarity to the source word. The presence of distractors, which may include translations of the target word, can lead to non-unique solutions for word translation tasks. To minimize ambiguity, we formulate multiple-choice questions and evaluate DeepSeek's generated options against the correct translations. Finally, we only retain the data in which the model's responses matched the target language word as our final dataset. 

\subsection{Analysis}

\begin{figure*}[t]
  \includegraphics[width=0.28\linewidth]{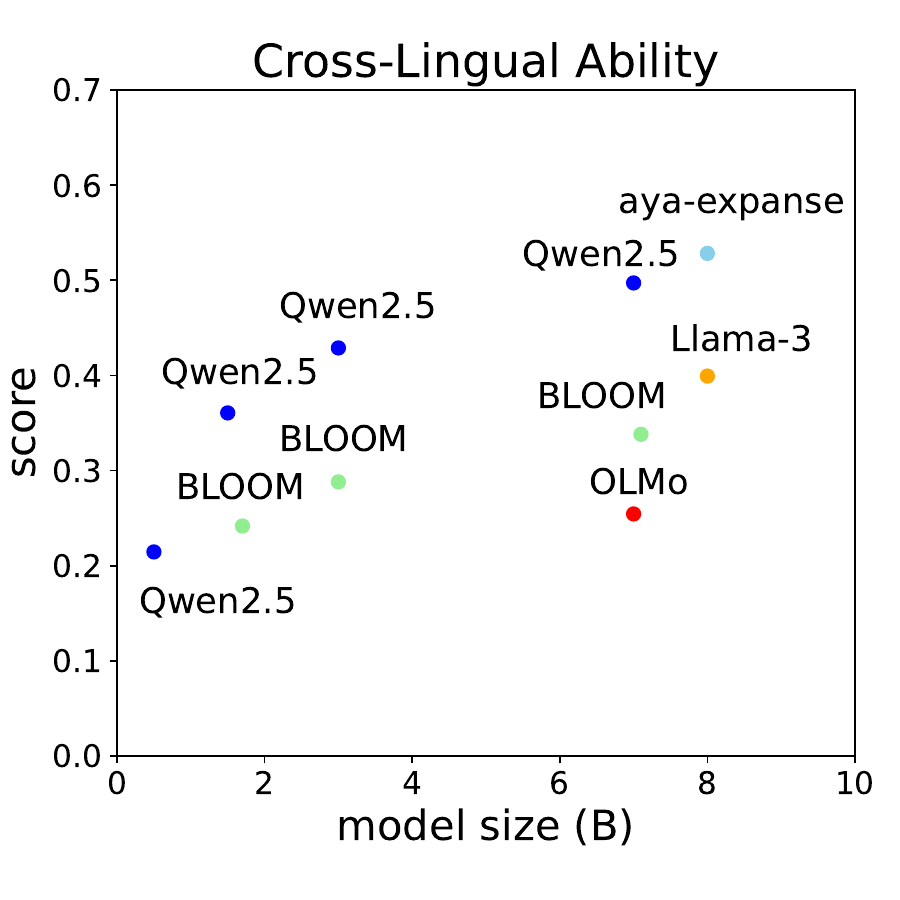}
  \hfill
  \includegraphics[width=0.28\linewidth]{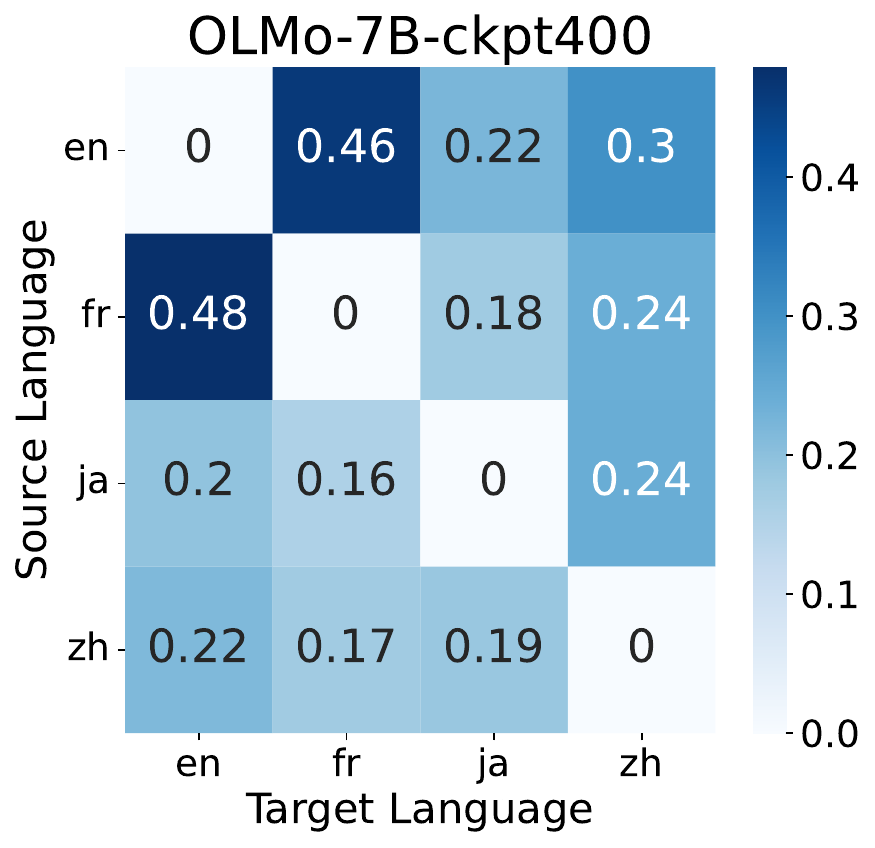}
   \includegraphics[width=0.42\linewidth]{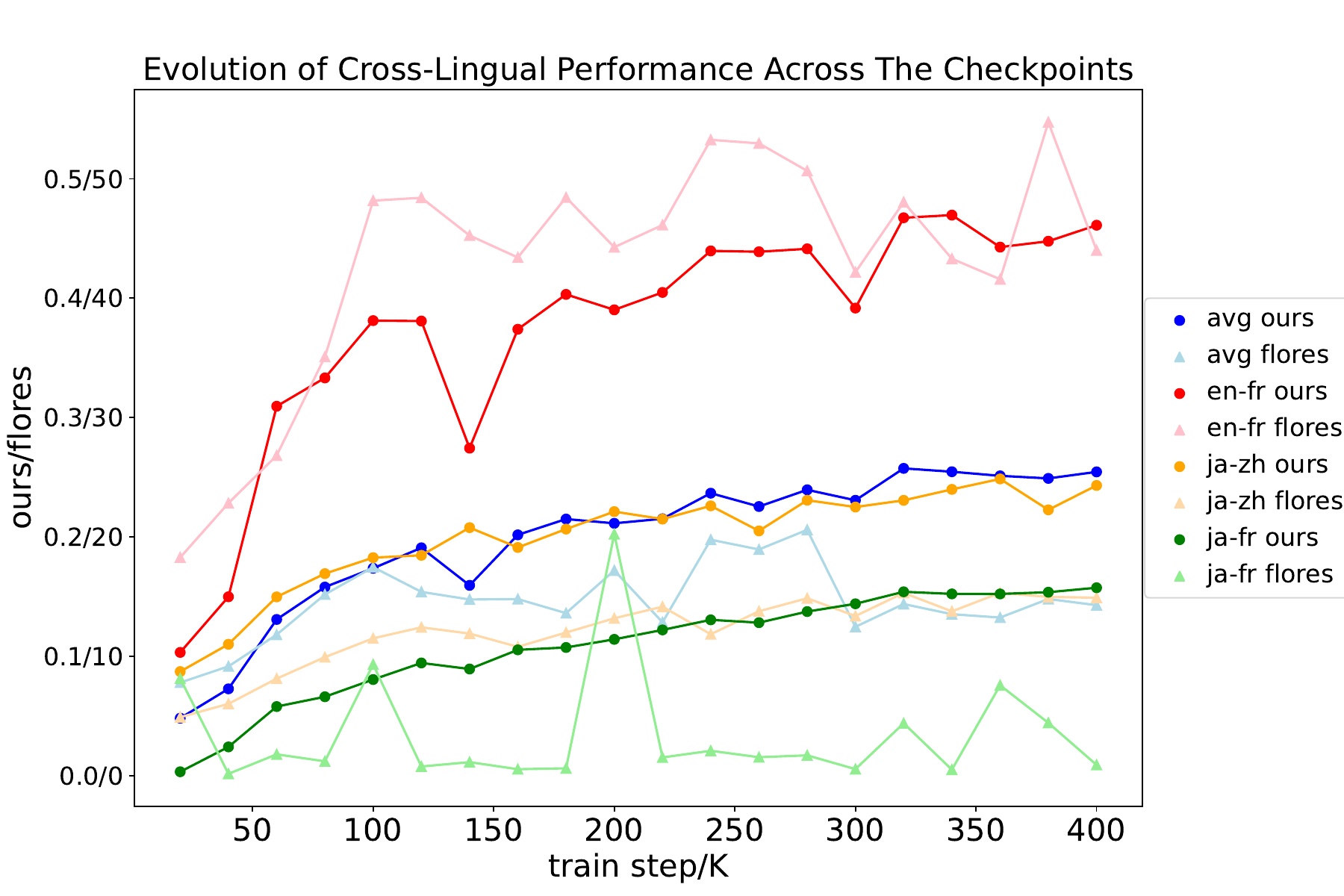}
  \caption{Different series of models' distribution of cross-language ability scores shown in \textbf{left}. Models belonging to the same family are consistently represented by the same color. The specific distribution of the OLMo-7B-0424 model's cross-lingual ability score matrix is shown in \textbf{middle}. The changes in our metric and flores scores during the training process of the OLMo-7B are shown in \textbf{right}. Different shapes are used to distinguish different tasks, and different colors are used to distinguish different language pairs. "avg" represents the cross-lingual ability averaged across all language pairs.}
  \label{fig:model_score}
\end{figure*}

We evaluate the cross-lingual ability of several large language models, including the BLOOM series with 560M, 1.7B, 3B, and 7.1B parameters \citep{le2023bloom}, the Qwen2.5 series with 0.5B, 1.5B, 3B, and 7B parameters \citep{qwen2.5}, the Llama-3 model with 8B parameters \citep{llama3modelcard}, the OLMo series with 7B parameters, step 400k \citep{OLMo}, and the Aya Expanse model with 8B parameters \citep{dang2024ayaexpansecombiningresearch}. The results of the cross-lingual evaluation are presented in Figure \ref{fig:model_score}. The cross-lingual ability score matrices of the remaining models are presented in Appendix \ref{a}. 
 
With an increase in model size, the cross-language ability continuously increases. The OLMo-7B model, trained on the Dolma Dataset - a pre-training dataset predominantly in English with minor documents in other languages, shows better cross-lingual ability in English-French. Meanwhile, the cross-lingual score matrix exhibits good symmetry, enabling the cross-lingual ability scores can be compared both horizontally and vertically. Aya Expanse 8B, a model known for cross-lingual ability, also demonstrates more powerful capabilities on this dataset. This also further validates the effectiveness of our Cross-Lingual Word Translation Task. To demonstrate the superiority of our method in evaluating cross-lingual model capabilities, we compute scores for ours and chrF++ on flores across OLMo-7B models' checkpoints and present results for language pairs representing low, medium, and high levels of cross-lingual abilities. Our metric demonstrates a more gradual increase and a more obvious change, particularly in language pairs with lower cross-lingual scores (e.g., Japanese-French). This implies that it can be used to compare models with limited cross-lingual abilities.

\section{Semantic Pivot behavior} 
\label{sec:section4}

Previous works \citep{wendler2024llamas, zhong2024beyond} analyzing intermediate layer outputs only focus on the language of the outputs,  neglecting the crucial relationship between these outputs and the model's input and final output. To investigate the source of cross-lingual ability and inference mechanisms in LLMs, we discover and distinguish LLMs' two distinct behaviors when LLMs do word translation tasks.

\subsection{Explore Potential Mode of Inference}
\label{sec4.1}
\begin{figure}[t]
    \centering
    \includegraphics[width=\columnwidth]{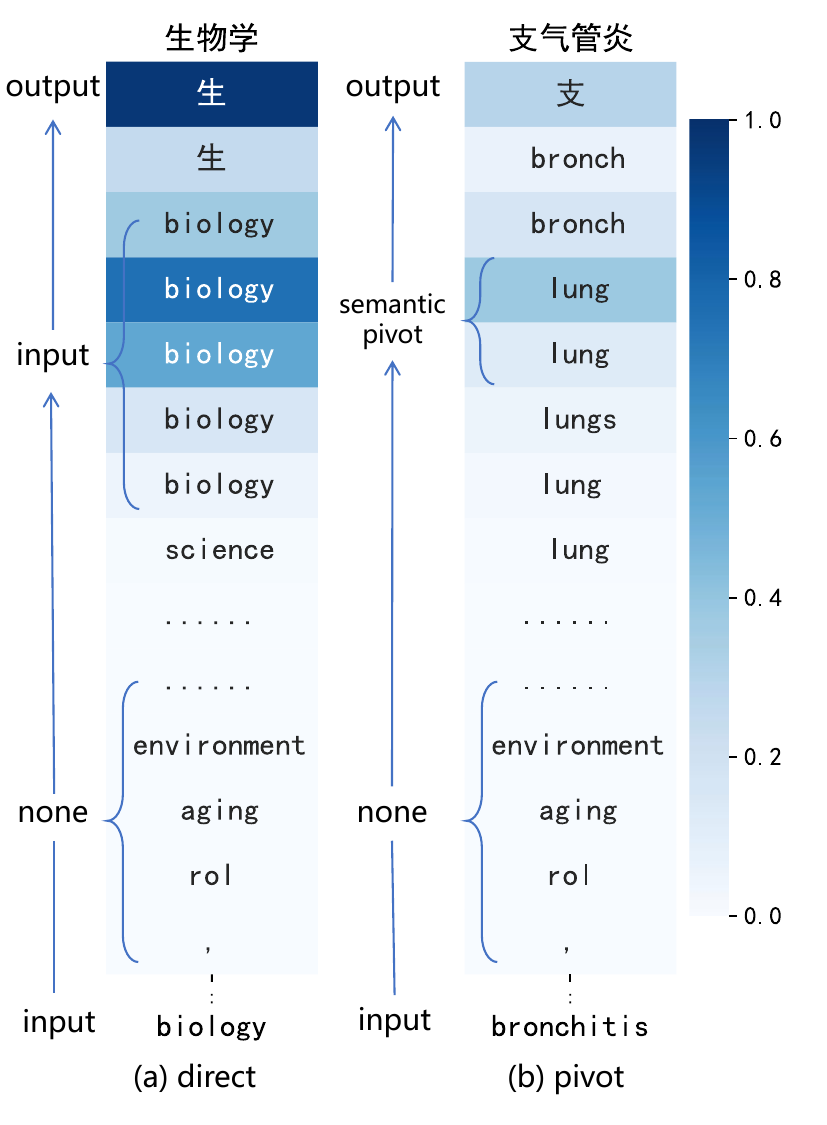}
    \caption{The example of the model's logit lens at the first output token in 31 layers. The \textbf{left} side illustrates the model's forward pass process of the co-occurrence behavior, while the \textbf{right} side depicts the process of the semantic pivot behavior.}
    \label{fig:example}
\end{figure}

To facilitate the interpretation of the model's internal mechanisms, we apply the \textbf{logit lens}, which can decode the model's hidden output and provide tokens' probability distribution in intermediate layers, and summarize them into two behaviors. To mitigate concerns that the token is a fluke of the model's internal representation, we mainly focus on the result of the token with the highest output probability in all layers as the semantic pivot.

\begin{itemize}
    \item \textbf{Co-occurrence behavior:} LLM directly infers the output from the input (input $\to$ output). 
    \item \textbf{Semantic pivot behavior:} LLM firstly obtains the semantic pivot in the intermediate layer from the input and then infers the output from the semantic pivot (input $\to$ semantic pivot $\to$ output).
\end{itemize}

We present the logit lens of the OLMo model performing the word translation tasks (e.g., English to Chinese) and decode the tokens into words in Figure~\ref {fig:example}.  When the OLMo model translates "biology" from English to Chinese as shown in \textbf{a}, the source word directly does the forward pass to the target word between layers. This demonstrates that the model employs the co-occurrence behavior during translation. When LLM translates "bronchitis" from English to Chinese as shown in \textbf{b}, the token corresponding to the word "lung" has the highest probability in all layers. LLM utilizes this token for transition at the fourth layer from the bottom, demonstrating that LLM reasons by leveraging semantic pivots.

\subsection{Distinguish Between the Two Behaviors}
\label{sec4.2}
To determine the behaviors adopted by models during inference, we establish a criterion based on a token with the highest probability across all layers to distinguish between the two behaviors. Specifically, if this token is present in either the inputs or the outputs, we classify the model's inference as the co-occurrence behavior. Otherwise, we categorize it as the semantic pivot behavior.

\begin{table}
  \centering
  \begin{tabular}{l|cccc}
    \hline
     & en & fr & zh & ja\\
    \hline
    en &      & 0.64 & 0.65 & 0.62 \\
    fr & 0.63 &      & 0.58 & 0.54 \\
    zh & 0.62 & 0.54 &      & 0.49 \\
    ja & 0.60 & 0.52 & 0.51 &      \\
    \hline
  \end{tabular}
  \caption{AUC score on each language pair using source word and target word co-occurrence frequency. Source languages are listed in the first column and target languages are listed in the first row.}
  \label{tab:AUC}
\end{table}

To understand the underlying learning mechanisms that drive these two behaviors, we suppose that high-frequency co-occurring words tend to exhibit co-occurrence behavior, while models for low-frequency contributing words rely on semantic pivots. Thus, we conduct a co-occurrence frequency analysis. We use WIMBD \citep{elazar2023s}, a set of tools for analyzing and revealing the content of large-scale datasets, to get the number of documents containing the source word and the target word in the Dolma dataset. Recognizing that identical strings for source and target words preclude accurate co-occurrence counts, we exclude instances where the source and target words have the same string. We identify data that satisfy the co-occurrence behavior as positive examples and calculate the AUC score, a standard metric for binary classification tasks. The AUC score can be expressed as:
$$P = \{x \mid x \in \text{semantic pivot behavior}\}$$
$$D = \{x \mid x \in \text{direct behavior}\}$$
$$AUC=\frac{\sum_{x\in D}rank(x)-\frac{|D|(|D| + 1)}{2}}{|D||P|}$$
$|P|$ and $|D|$ represent the number of data inferred by LLM via semantic pivot behavior and direct behavior. $rank(x)$ presents the rank of the number of documents containing both the source word and the target word for data $x$ among all datasets.

The result is shown in Table ~\ref{tab:AUC}. The majority of AUC scores exceed $0.5$, meaning that the co-occurrence behavior is more prevalent in translation tasks where the source and target words frequently co-occur, while semantic pivot behavior is more characteristic of translation tasks with low-frequency co-occurrence. Co-occurrence frequency in the three non-English languages doesn't exhibit strong discriminative capability in the three non-English low-resource languages.

\subsection{Find Semantic Pivots From Dataset}
\label{sec4.3}
To explore how models learn cross-lingual abilities through semantic pivots, we conduct a \textbf{token co-occurrence proportion analysis}. We suppose that the semantic pivot may have a high relative co-occurrence proportion with both the source word and the target word. To accurately quantify co-occurrence proportion, we select the translation task data where the source word and target word have no overlapping tokens and satisfy semantic pivot behavior. Next, for each word translation task, we independently sample 2,000 documents from the documents containing the source language word, denoted as $\text{doc}_s$, and the documents containing the target language word, denoted as $\text{doc}_t$ using Infini-gram API \citet{Liu2024InfiniGram}. Then, we count the frequency of tokens in these documents to estimate the co-occurrence proportion of all tokens with the source word and the target word throughout the entire pre-training process. For some high-frequency tokens, due to their widespread occurrence in multiple documents, co-occurrence frequency fails to capture their unique characteristic. To eliminate the impact of the high-frequency occurrence of tokens in the background on the search for semantic pivots, we subtract the token's frequency in the background $\text{Fre}$ from the frequency of its co-occurrence with the source word $\text{Fre}_{x_s}$ and the target word $\text{Fre}_{x_t}$. We select the smaller of these two adjusted relative frequencies $F(x)$ as the basis for ranking. 

$$F(x) = \min \big( \text{Fre}_{x_s} - \text{Fre}, \text{Fre}_{x_t} - \text{Fre} \big)$$

We select the top 50 as candidate semantic pivots and prompt Deepseek to filter out meaningless tokens and retain others as a potential set of semantic pivots. 

$$S_{\text{pivot}} = \text{Filter}_{LLM} \left( \text{Top}_{50} \left( \left\{ x \mid F(x)\right\} \right) \right)$$

\begin{figure}
    \centering
\includegraphics[width=0.80\columnwidth]{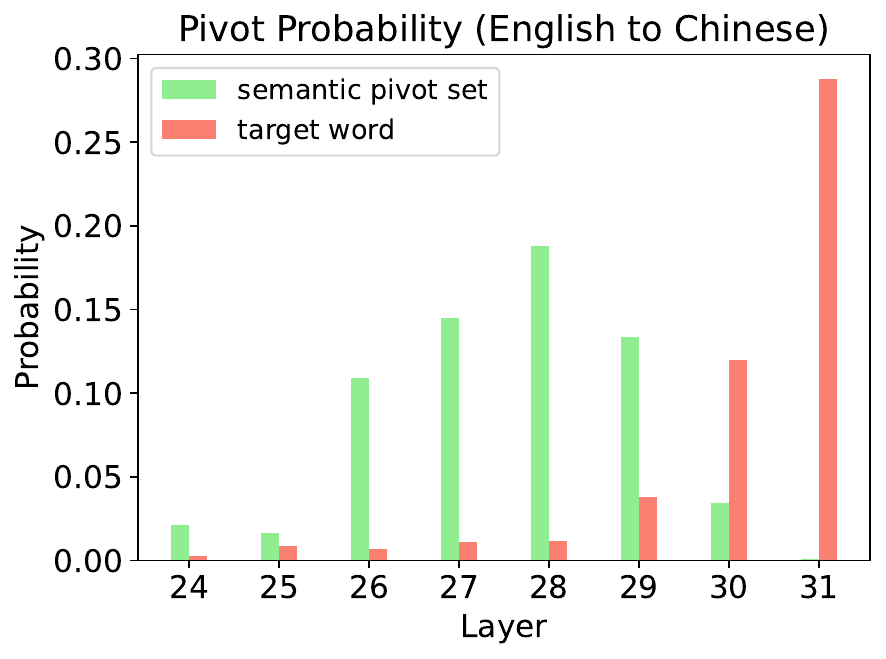}
    \caption{The semantic pivots set of probability in the last eight layers. The x-axis represents the OLMo-7B's layer index, and the y-axis indicates the total probability of all tokens in the semantic pivot set and the target word's tokens. }
    \label{fig:distribution}
\end{figure}

In order to verify the rationality of the semantic pivots we find, we further calculate the total probability of all semantic pivots in our set in each layer as shown in Figure \ref{fig:distribution} (e.g., English to Chinese). The probability of the semantic pivot set exhibits a trend of initially increasing, peaking at the 28th layer, and subsequently decreasing. This phenomenon suggests that during the translation process, the intermediate layer states of the model, to some extent, first transfer to the semantic pivots we are looking for, and then transform into the final outputs. Although the probability in each layer is not more than $0.2$, it significantly surpasses the average probability of a 50-token set, which is approximately $\frac{50}{50253}$, indicating the set of semantic pivots we find has a higher proportion compared to other tokens.
  
\section{Application}
\label{sec5}
Even though we have gained insights into how the model learns semantic pivots from the training dataset and applies them to perform word translation tasks, how to apply them to improve the cross-lingual ability score remains a challenge. Based on findings in the semantic pivot behavior, we propose a method for constructing a semantic pivot-aware pre-training dataset to enhance the model's cross-lingual ability. 

\subsection{How to Find Semantic Pivot in Pre-training Dataset}

\begin{figure*}[t]
  \includegraphics[width=0.90\linewidth]{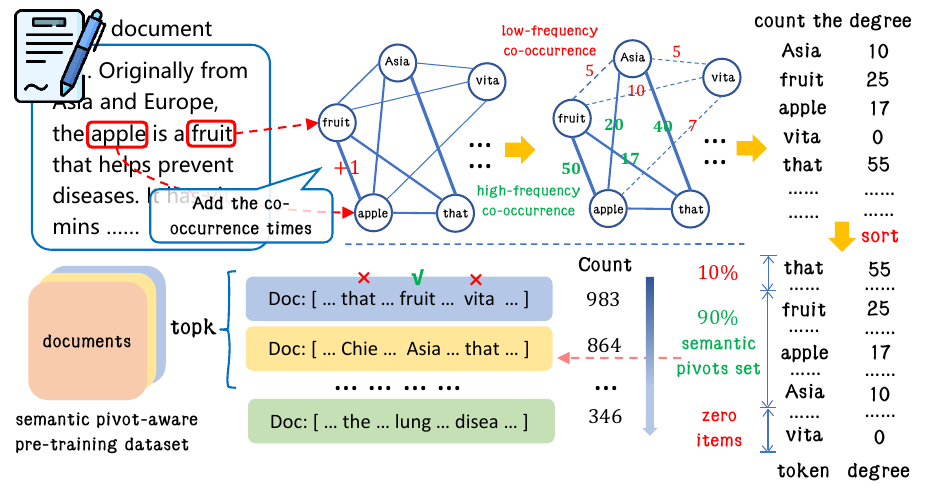}
  \caption{The process of constructing a semantic pivot-aware pre-training dataset used to improve the model's cross-lingual ability.}
  \label{fig:pretrain}
\end{figure*}

\label{sec5.1}
Based on the discovery in Section \ref{sec:section4}, the semantic pivots tend to exhibit a high relative co-occurrence proportion with both the source and target words, compared to the other tokens. Therefore, we introduce a \textbf{semantic pivot document purification} method to filter out tokens of both types with low relative co-occurrence frequency. 

We construct an adjacency matrix from the pre-training data to count the co-occurrence times between tokens. We set a frequency threshold as $10\%$ of the average probability of token occurrence within a single document to determine whether two tokens have a high-frequency co-occurrence. Then, for each token, we count the number of tokens that have a high co-occurrence frequency with it. For any token with no high-frequency co-occurring counterpart, we categorize it as a low-frequency token with a low relative co-occurrence frequency and filter it out. We sort the remaining tokens according to the number of their high co-occurrence frequency tokens. We consider the top $10\%$ of tokens as high-frequency tokens with low relative co-occurrence frequency and filter them out. We regard the remaining tokens as our semantic pivot set.
Finally, we compute the number of tokens belonging to the semantic pivot set within each document and select the top-ranked documents as a high semantic pivot proportion document to construct our pre-training dataset. We illustrate the process of constructing this pre-training data in Figure \ref{fig:pretrain}.

\subsection{Experiments}
\label{sec5.2}
\begin{table}
  \centering
  \begin{tabular}{l|cccc}
    \hline
     & fr & zh & ja & doc\\
    \hline
    Original & 0.1\% & 0.3\% & 0.2\% & 34k \\
    Multilingual & \textbf{1.7\%} & 4.3\% & \textbf{2.5\%} & \textbf{502k} \\
    Ours & 1.0\% & \textbf{6.0\%} & 1.5\% & 450k \\
    \hline
  \end{tabular}
  \caption{The distribution of the language in three types of datasets and the number of documents containing one of three non-English languages.}
  \label{tab:lang}
\end{table}

To validate the effectiveness of our proposed method for enhancing cross-lingual ability, we construct two datasets as baselines: an Original dataset and a Multilingual dataset. We train and evaluate the model using two baseline datasets and our dataset, constructed by the method in Section \ref{sec5.1}.

For the Original dataset, we duplicated the pre-training data from the corresponding checkpoints of the original model. For the multilingual dataset, we extract training documents in the three languages, French, Chinese, and Japanese, from the pre-training dataset of 15 checkpoints, following the corresponding checkpoint. For the remaining documents, we use the English documents from the original checkpoint for padding. For our dataset, we extract training documents in French, Chinese, and Japanese from the pre-training data of 25 checkpoints, as multilingual candidate documents. We construct a cross-lingual adjacency matrix on a subset of multilingual documents and employ our method to rank the multilingual candidate documents. We selected the top-ranked approximately 450,000 documents for our dataset, ensuring the aggregate proportion of low-resource languages aligns with that of multilingual datasets.

To achieve more accurate language identification and filtering of documents, we divided each document into sliding chunks with a chunk size of 256 and a sliding step of 128. Then we utilize fasttext \citep{joulin2016fasttext} to identify the language of each chunk and select the result with the highest probability as the language of that chunk. The distribution details of languages in the three datasets are presented in Table \ref{tab:lang}. Currently, there is a scarcity of publicly available multilingual large language models with open-source checkpoint parameters and pre-training data. We choose the OLMo model with a size of 1B, a model trained on the open-source pre-training dataset Dolma, and publish the training data of each step. In order to better replicate the original pre-training process and conduct a fair comparison, we utilize the training code and parameters provided by allenai \citep{OLMo}. We separately used the three datasets to continue the training for 1000 steps at the 605th checkpoint. This checkpoint is located at the late of the training process, where the model's performance change remains relatively stable after training. 

\begin{table}[t]
    \centering
    \resizebox{0.90\columnwidth}{!}{
    \begin{tabular}{cc|cccc}
        \toprule
        \multirow{2}{*}{\textbf{Setting}} & \multirow{2}{*}{} & \multicolumn{4}{c}{\textbf{Target Language}}  \\
        \cmidrule{3-6}
        & & en & fr & zh & ja \\
        \midrule
        \multirow{4}{*}{\shortstack{Original\\dataset\\avg: 0.1420}}
            & en &      & 0.31 & 0.13 & 0.18 \\
            & fr & 0.34 &      & 0.08 & 0.11 \\
            & zh & 0.09 & 0.05 &      & 0.13 \\
            & ja & 0.11 & 0.06 & 0.12 &      \\
        \midrule
        \multirow{4}{*}{\shortstack{Multilingual \\ dataset\\avg: 0.1506}}
            & en &      & 0.33 & 0.13 & 0.19 \\
            & fr & 0.33 &      & 0.09 & 0.11 \\
            & zh & 0.09 & 0.06 &      & 0.15 \\
            & ja & 0.12 & 0.08 & 0.14 &      \\
        \midrule
        \multirow{4}{*}{\shortstack{Our dataset \\ avg: \textbf{0.1552}}}
            & en &      & 0.32 & 0.13 & 0.20 \\
            & fr & 0.36 &      & 0.09 & 0.12 \\
            & zh & 0.10 & 0.06 &      & 0.15 \\
            & ja & 0.14 & 0.08 & 0.13 &      \\
        \bottomrule
    \end{tabular}}

    \caption{The cross-lingual ability score matrices of the model continue training on three datasets, mentioned in Section \ref{sec5.2} }
    \label{tab:matrices}
\end{table}

\subsection{Analysis}

We evaluate models' cross-lingual ability after training on three datasets, showing the results of the cross-lingual ability matrices in Table \ref{tab:matrices}. The multilingual dataset has more multilingual documents than the original training data. The model trained on a multilingual dataset exhibits stronger cross-lingual ability across almost all language pairs than the original model, meaning that more multilingual documents can effectively enhance cross-lingual ability. The model trains on our dataset achieves a 0.013 improvement over the original dataset baseline and a 0.005 improvement over the multilingual dataset baseline, even though the multilingual proportion in this dataset aligns with that of a multilingual dataset. The results indicate that our method of filtering documents with a high proportion of semantic pivots and using them to construct pre-training data effectively and comprehensively improves cross-lingual capabilities.

\section{Conclusions}
In this paper, to quantify the cross-lingual ability of LLMs accurately, we propose a Word-Level Cross-lingual Ability Task and evaluate it on LLMs. To find how LLMs learn cross-lingual ability, we identify and distinguish LLMs' two behaviors. We attribute LLMs’ two behaviors to the co-occurrence frequency of strings and calculate AUC score. We use token co-occurrence proportion analysis to identify semantic pivots from the pre-training dataset.
Finally, to improve the model's cross-lingual ability based on our findings, we do semantic pivot document purification to reconstruct the pre-training dataset rather than incorporating domain-specific data. Our experiment effectively improves the model's cross-lingual ability compared to the two baselines. 

\section*{Limitations}

 Due to the limited availability of open-source large language models with public checkpoints of pre-training data, training parameters, and pre-training code, we only validate our method on the OLMo-1B model. We are unable to conduct our experiments on larger language models with stronger cross-lingual ability.

\bibliography{custom}

\appendix
\section{Dataset Composition}
\label{a}
In our work, we evaluate LLMs' cross-lingual ability on a word translation task using the CLWTD. We analyzed the part-of-speech distribution of all words in the CLWTD shown in Table \ref{tab:dataset1}.

\label{sec3}
\begin{table}
  \centering
  \begin{tabular}{l|cccc}
    \hline
    word type & n & v & adj & total\\
    \hline
    CLWTD & 1423 & 173 &  404 & 2000 \\
    \hline
  \end{tabular}
  \caption{The part-of-speech distribution of words in CLWTD used in our research}
  \label{tab:dataset1}
\end{table}

\section{Models' Cross-Lingual Ability}
\label{b}

\begin{figure*}[t]
  \includegraphics[width=0.33\linewidth]{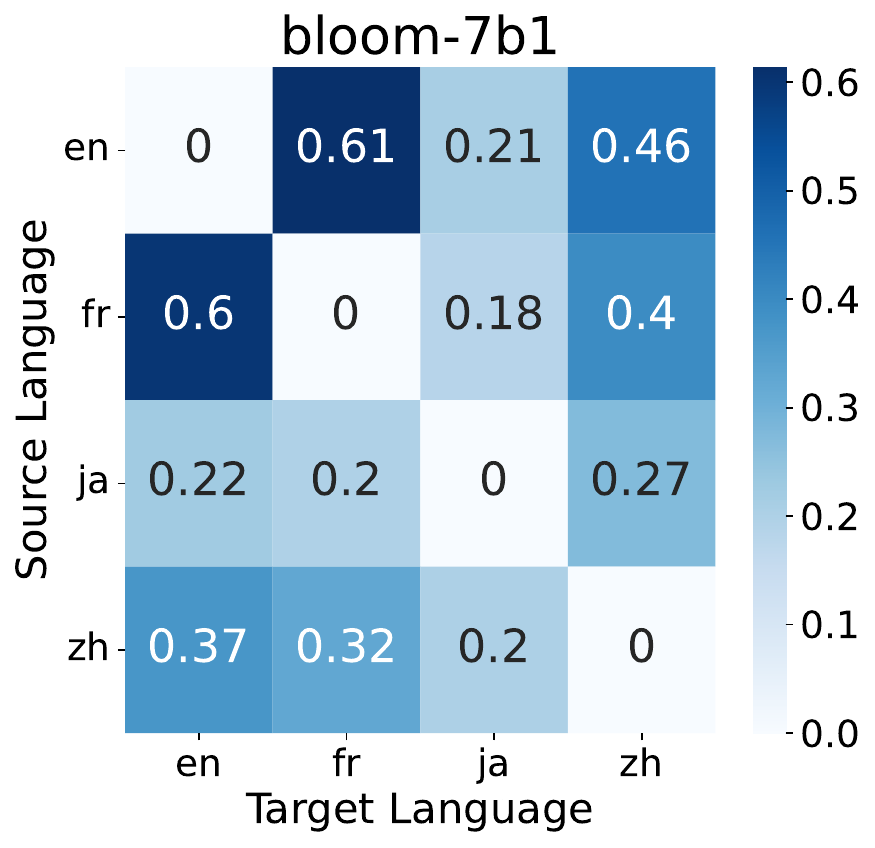}
  \includegraphics[width=0.33\linewidth]{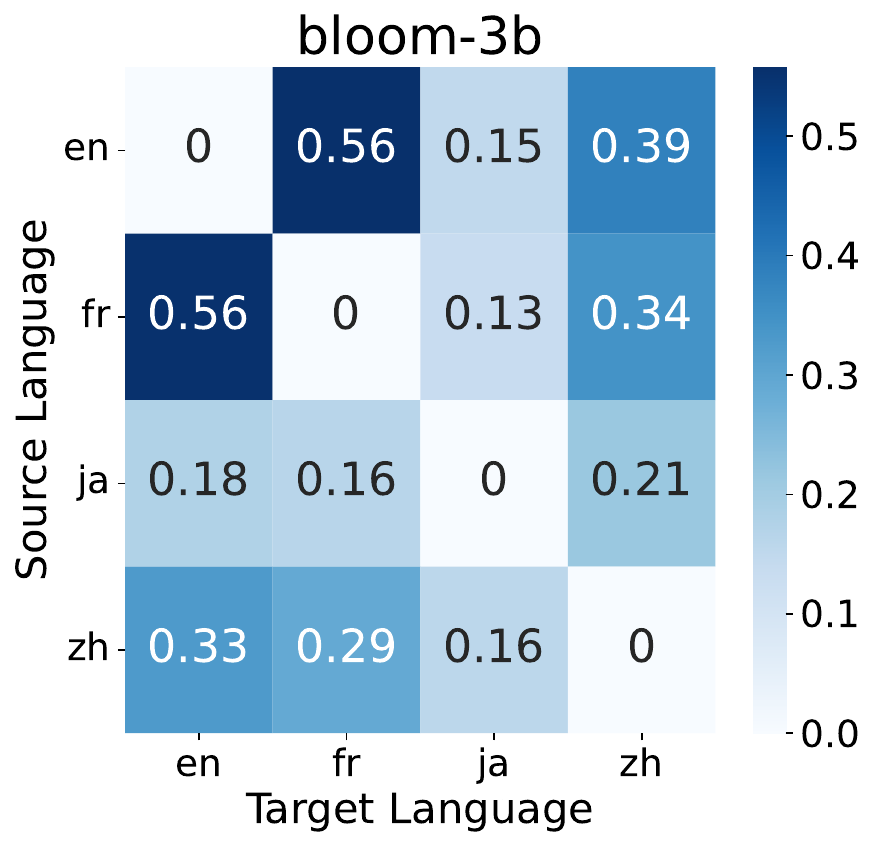}
  \includegraphics[width=0.33\linewidth]{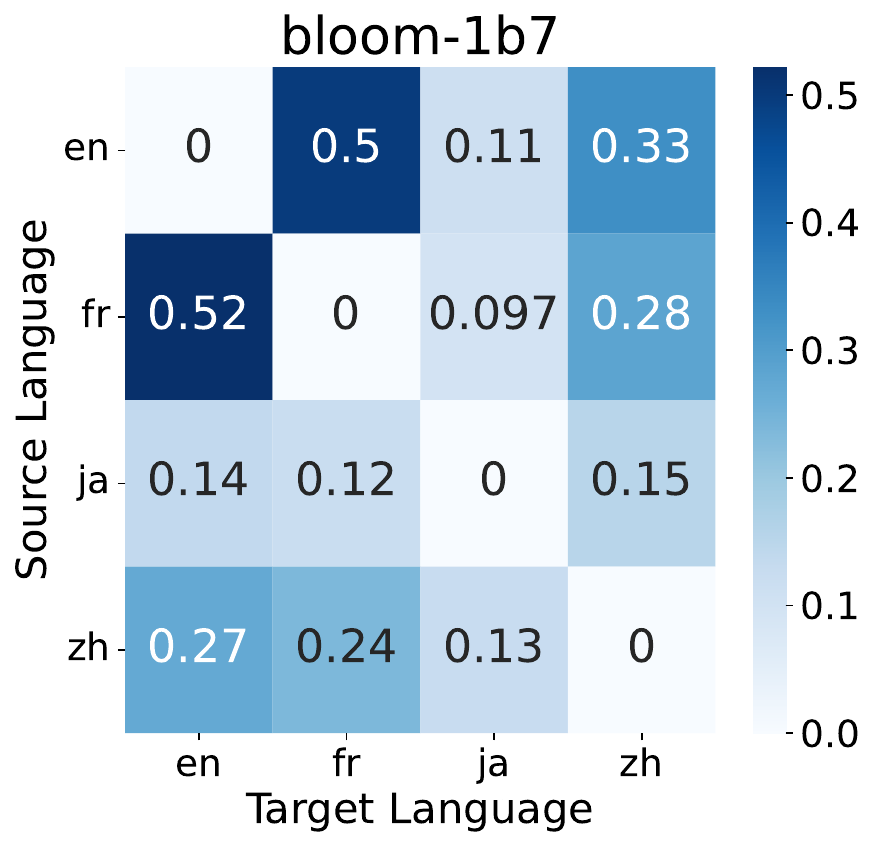}
  \includegraphics[width=0.33\linewidth]{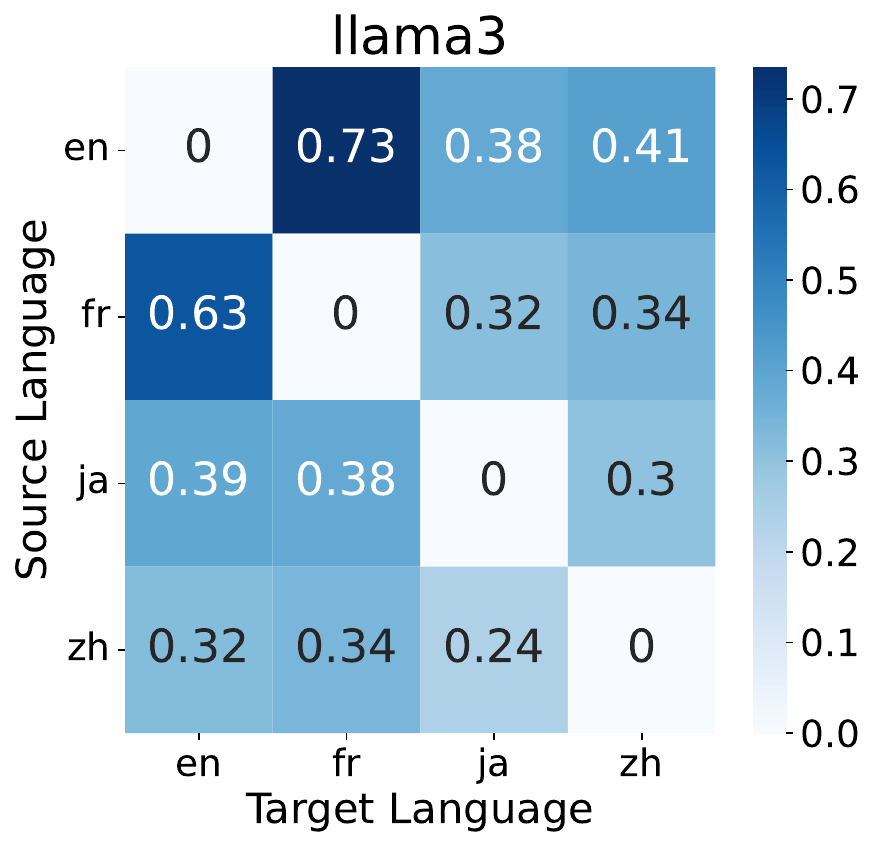}
  \includegraphics[width=0.33\linewidth]{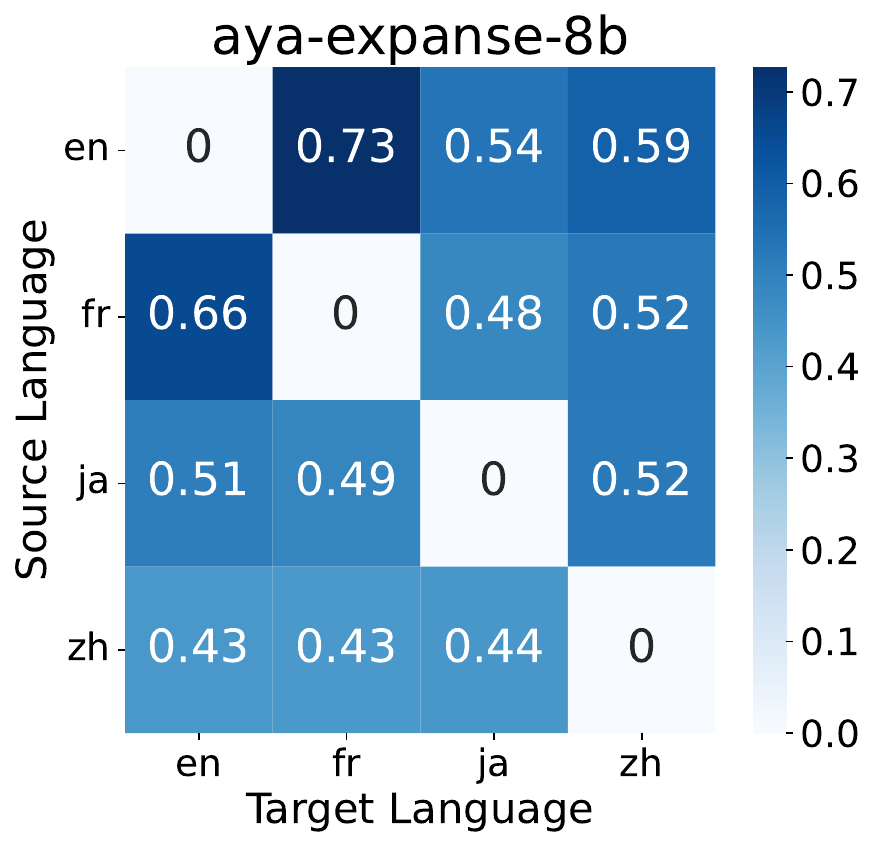}
  \includegraphics[width=0.33\linewidth]{latex/image/score/OLMo-7B-ckpt400.pdf}
  \includegraphics[width=0.24\linewidth]{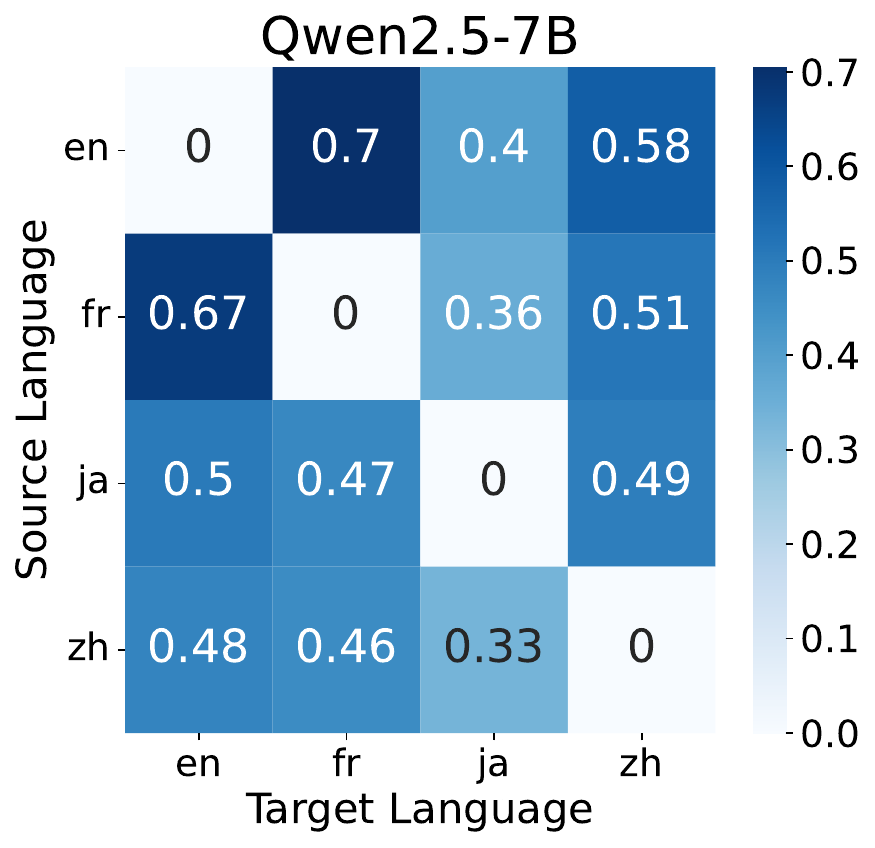}
  \includegraphics[width=0.24\linewidth]{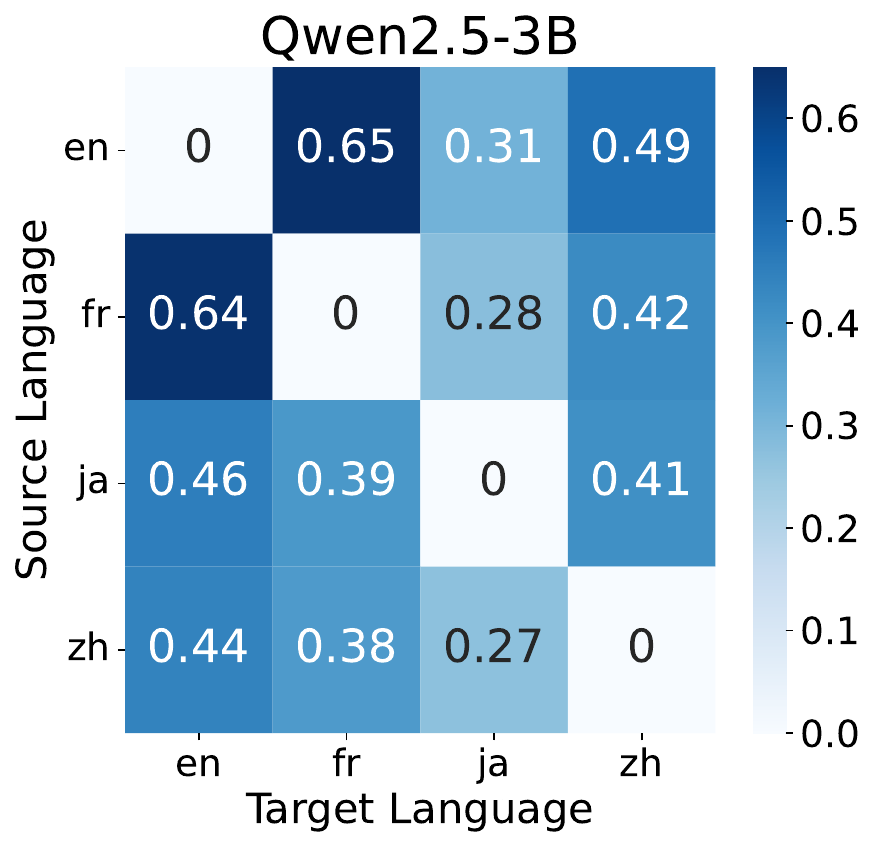}
  \includegraphics[width=0.24\linewidth]{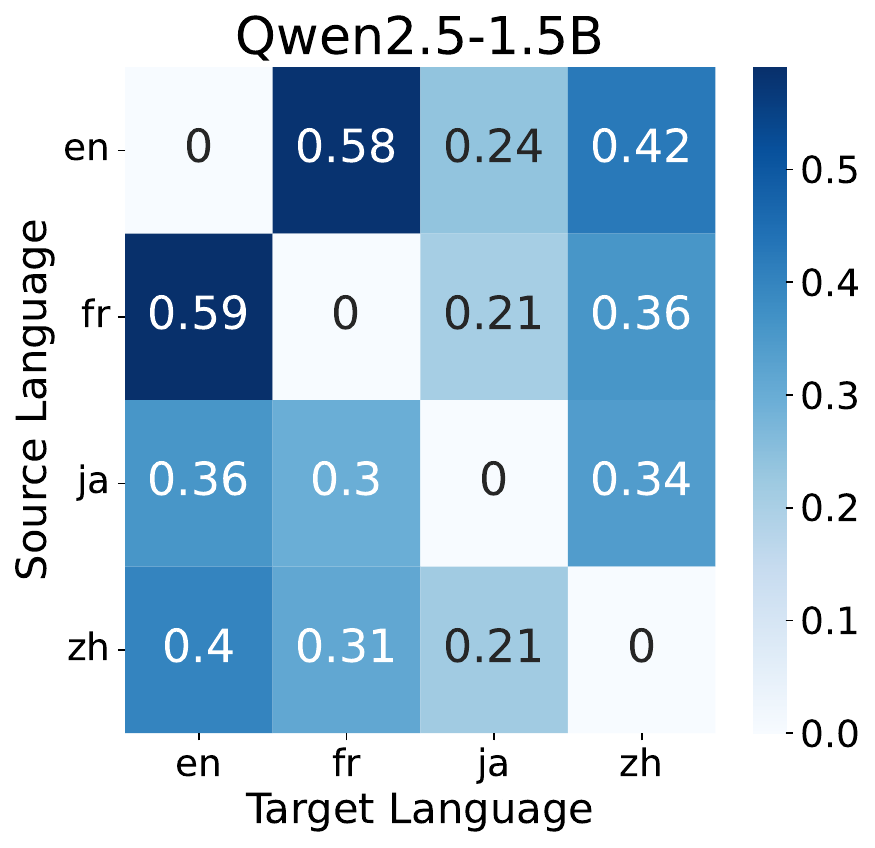}
  \includegraphics[width=0.24\linewidth]{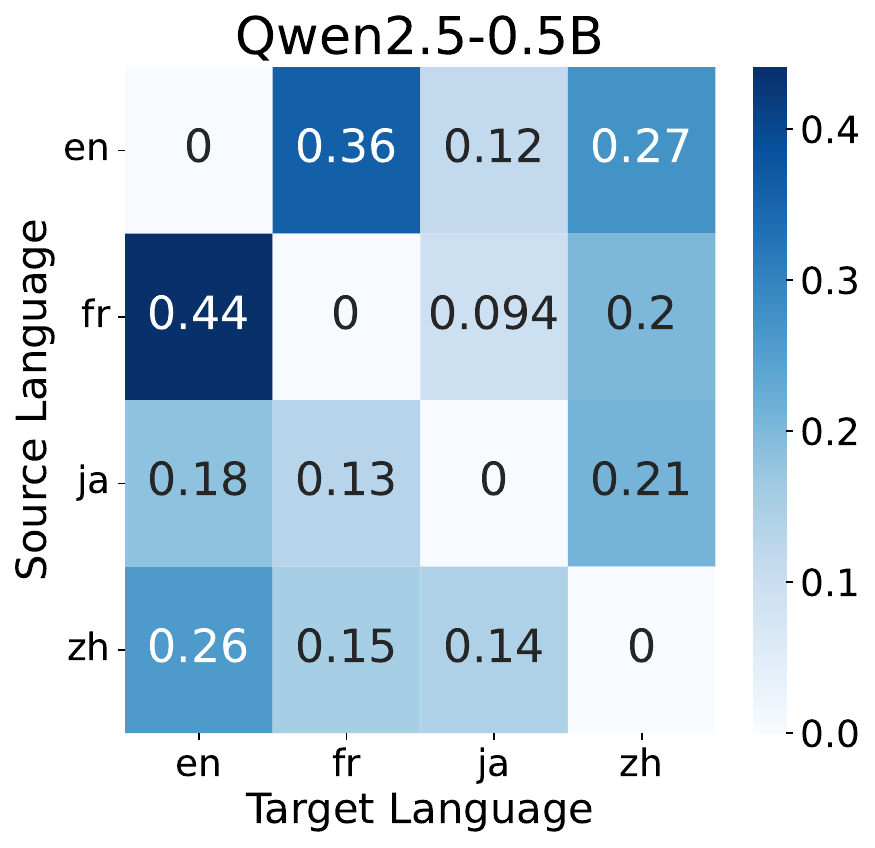}
  \caption{The specific distribution of the model's cross-lingual ability score matrix. The title describes the model we evaluate. The vertical axis represents the source language; the horizontal axis represents the target language.}
  \label{fig:all_score}
\end{figure*}

We provide details of the model's cross-lingual score matrix, we evaluate in Figure \ref{fig:model_score} shown in Figure \ref{fig:all_score}, including the BLOOM series with 560M, 1.7B, 3B, and 7.1B parameters, the Qwen2.5 series with 0.5B, 1.5B, 3B, and 7B parameters, the Llama-3 model with 8B parameters, the OLMo series with 7B parameters, step 400k, and the Aya Expanse model with 8B parameters. 
    
\begin{figure*}[t]
  \includegraphics[width=0.48\linewidth]{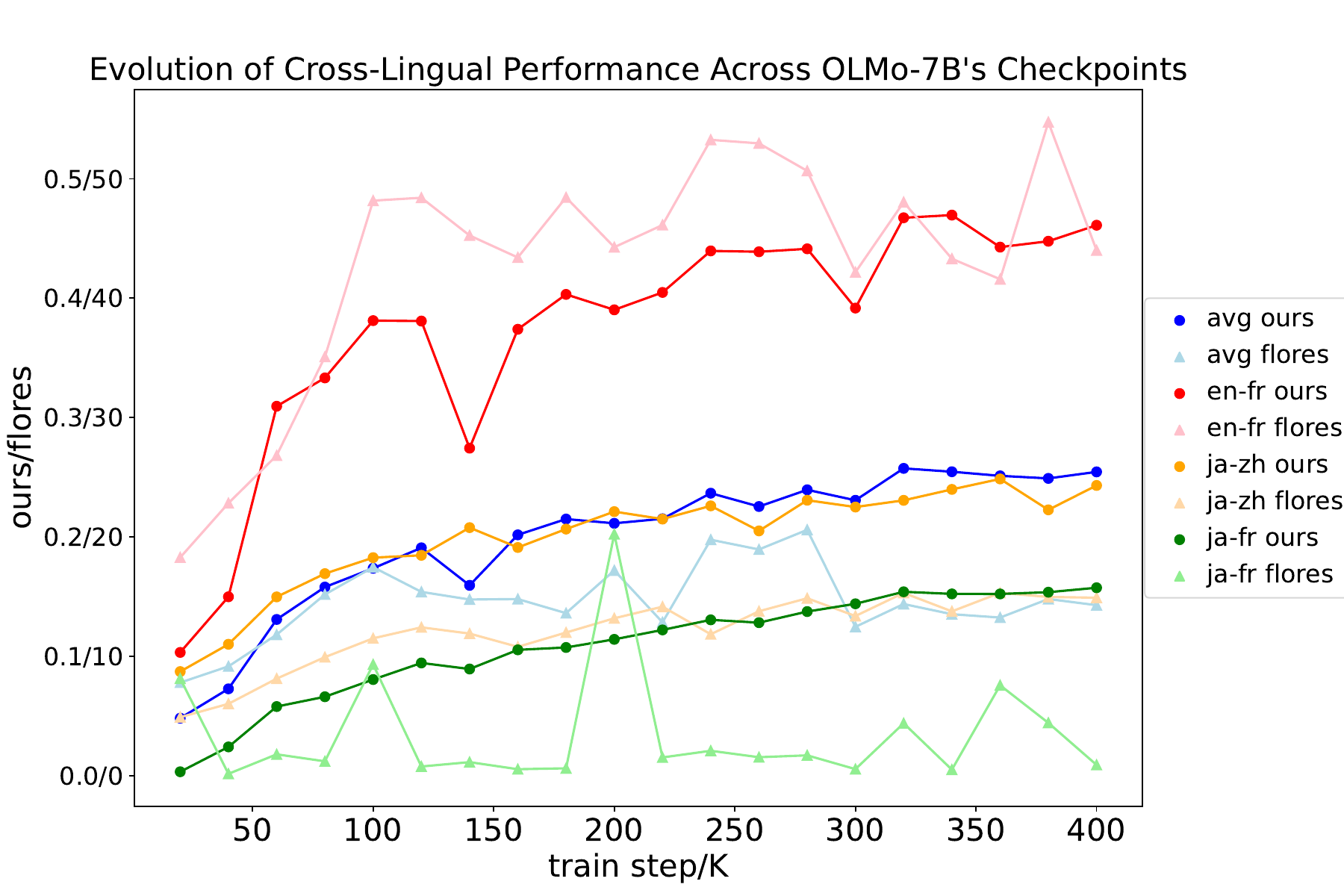}
  \hfill
   \includegraphics[width=0.48\linewidth]{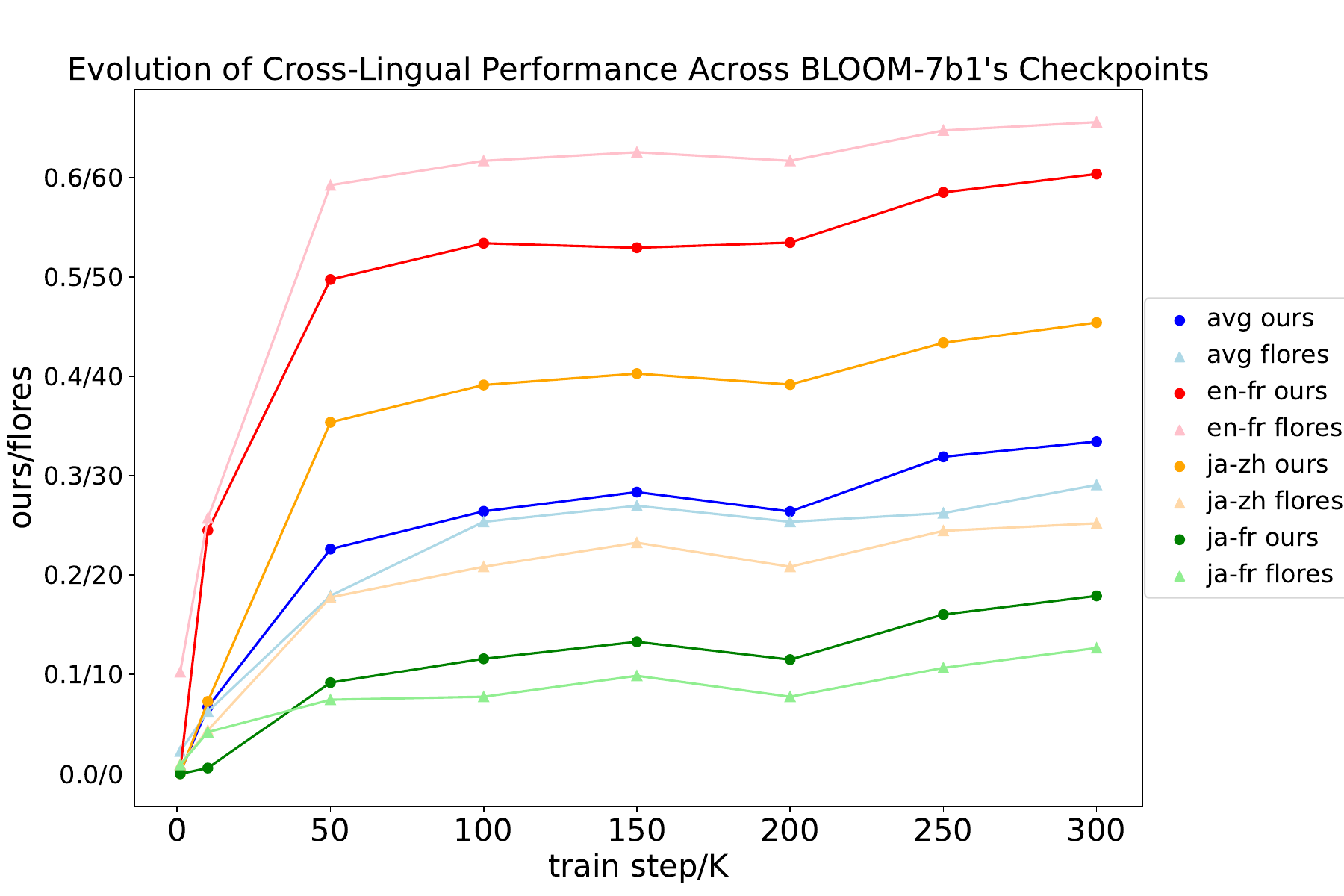}
  \caption{The changes in our metric and flores scores during the training process. The result of OLMo-7B are shown in \textbf{left}, and the result of bloom-7b1 are shown in \textbf{right} Different shapes are used to distinguish different tasks, and different colors are used to distinguish different language pairs. "avg" represents the cross-lingual ability averaged across all language pairs.}
  \label{fig:step_score}
\end{figure*}

To compare our method with the Sentence-level machine translation task, we compute scores for our proposed method and chrF++ on the FLORES dataset of models' checkpoints. We present results for language pairs representing low, medium and high levels of cross-lingual transfer ability in Figure \ref{fig:step_score}

\section{Prompt Design}
\begin{figure*}
    \centering
    \includegraphics[width=0.95\columnwidth]{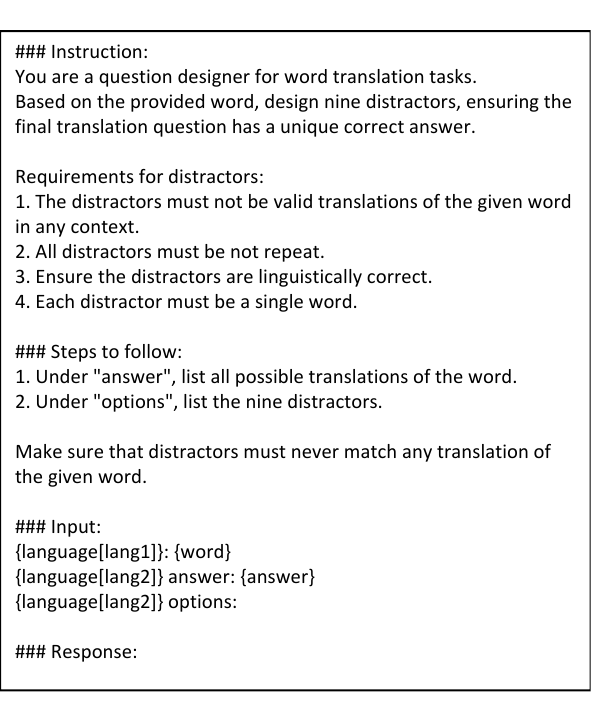}
    \caption{The prompt we use to generate the distractors}
    \label{fig:prompt2}
\end{figure*}

\begin{figure*}
    \centering
    \includegraphics[width=0.95\columnwidth]{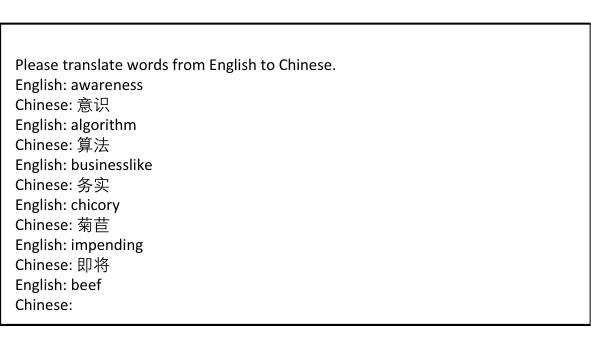}
    \caption{The example of a five-shot prompt used to calculate the model's cross-lingual ability}
    \label{fig:prompt}
\end{figure*}

\begin{figure*}
    \centering
    \includegraphics[width=0.95\columnwidth]{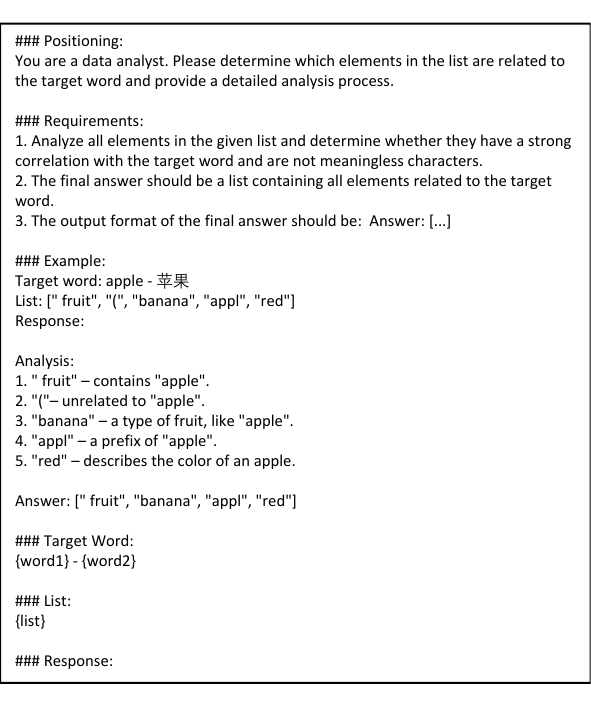}
    \caption{The prompt we design to filter out words irrelevant to the source word and the target word.}
    \label{fig:prompt3}
\end{figure*}

\label{c}
In Section \ref{3.1}, we prompt DeepSeek to generate distractors. we show the detail of the prompt in Figure \ref{fig:prompt2}.

We design the five-shot prompt template to calculate the model's cross-lingual ability. We show an example of a prompt requiring the model to translate the word "beef" from English to Chinese in Figure \ref{fig:prompt}.

To explore how models learn cross-lingual abilities through semantic pivots and discover the potential semantic pivot, we prompt Deepseek to filter out meaningless tokens. The  prompt is shown in Figure \ref{fig:prompt3}

\section{Result of Finding Semantic Pivots}
\begin{figure*}[t]
  \includegraphics[width=0.33\linewidth]{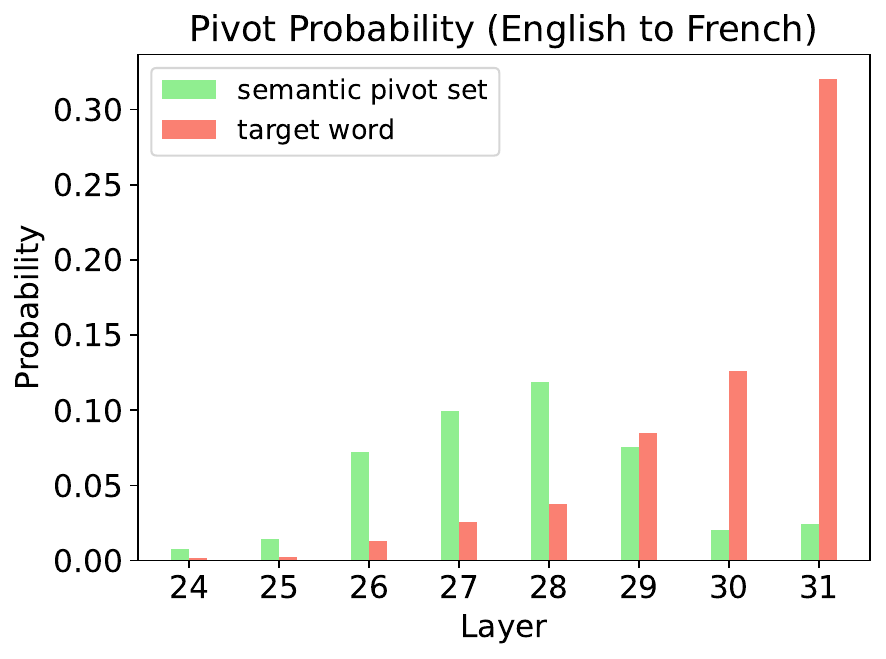}
  \includegraphics[width=0.33\linewidth]{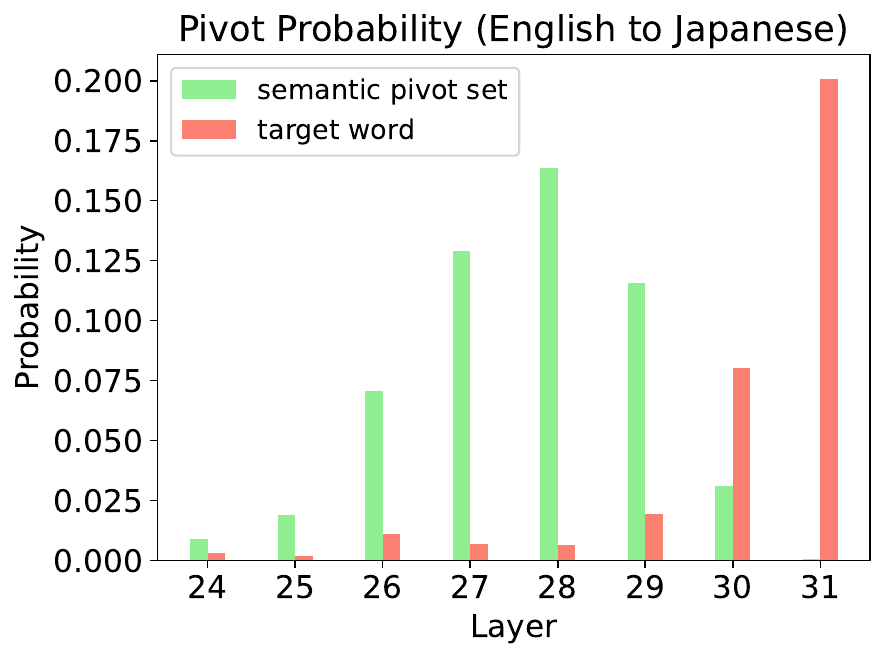}
  \includegraphics[width=0.33\linewidth]{latex/image/distribution/en-zh.pdf}
   \includegraphics[width=0.33\linewidth]{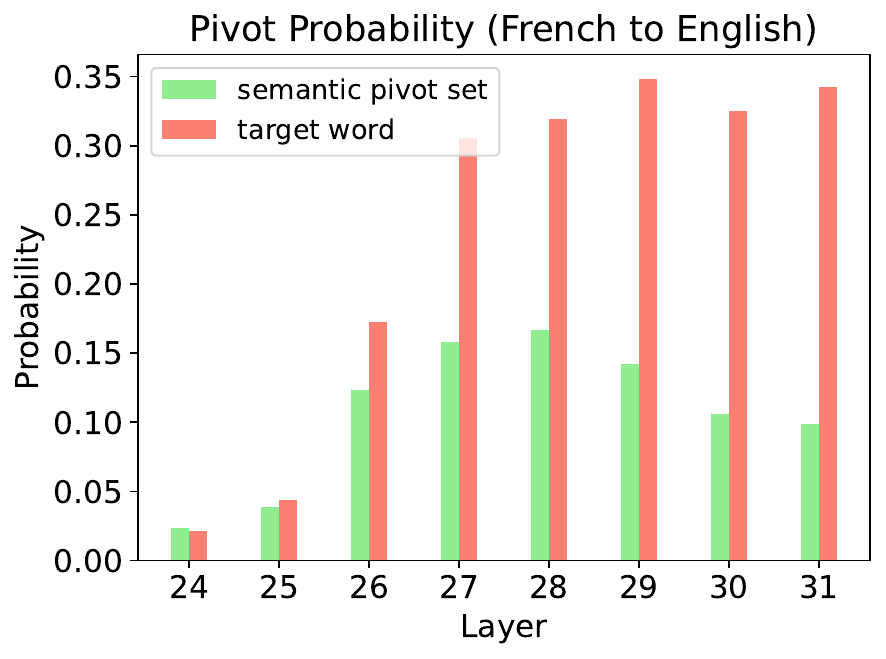}
  \includegraphics[width=0.33\linewidth]{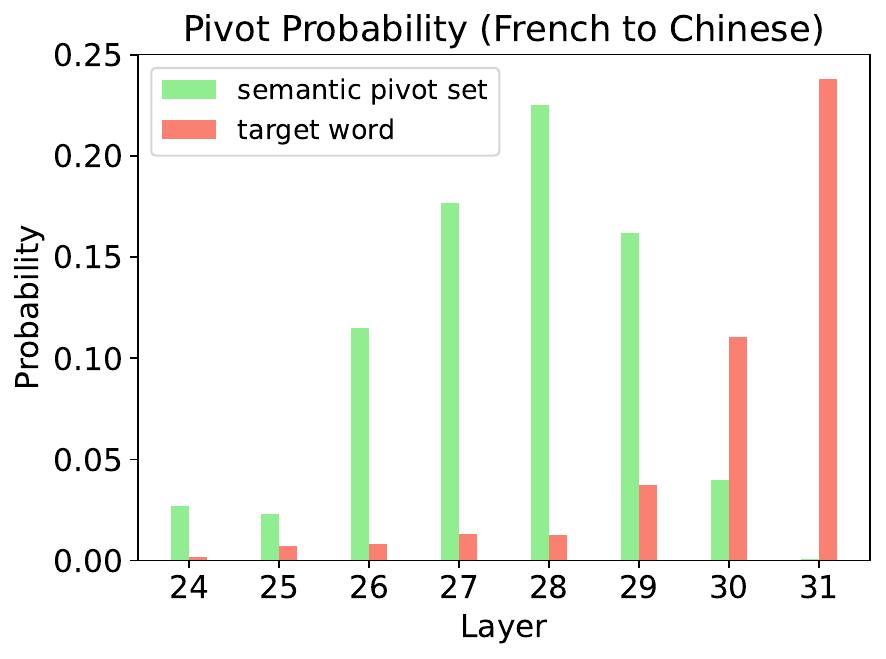}
  \includegraphics[width=0.33\linewidth]{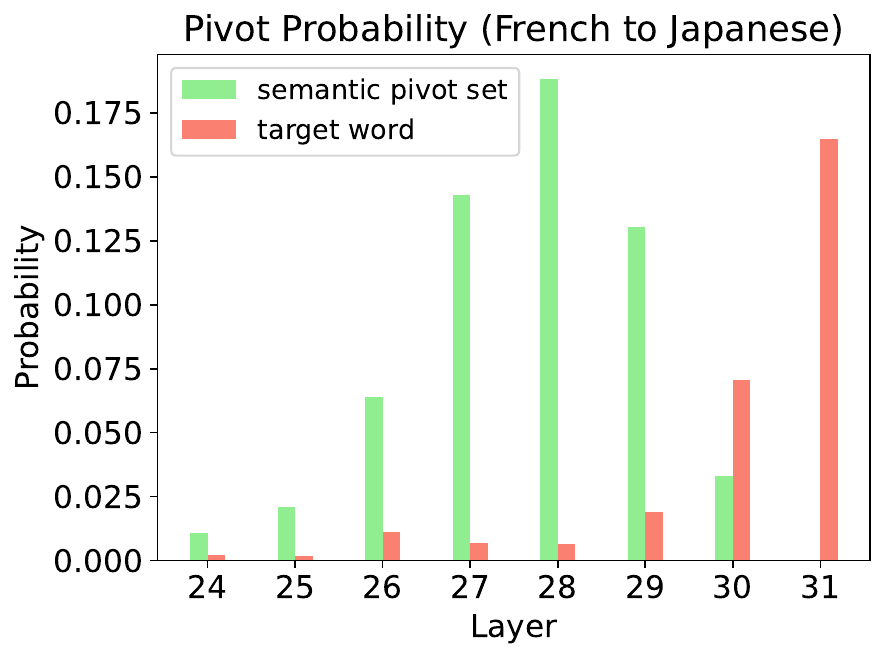}
    \includegraphics[width=0.33\linewidth]{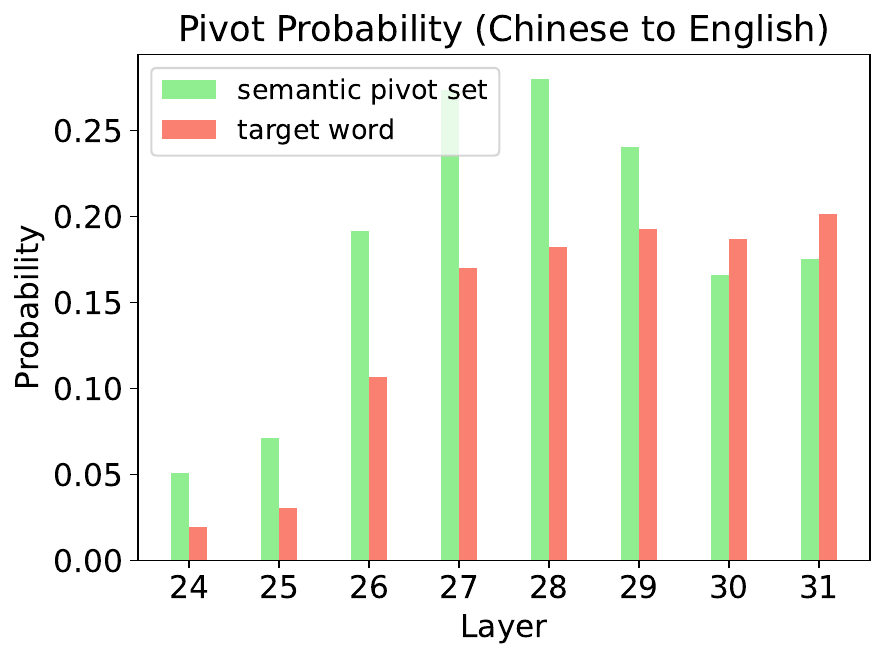}
  \includegraphics[width=0.33\linewidth]{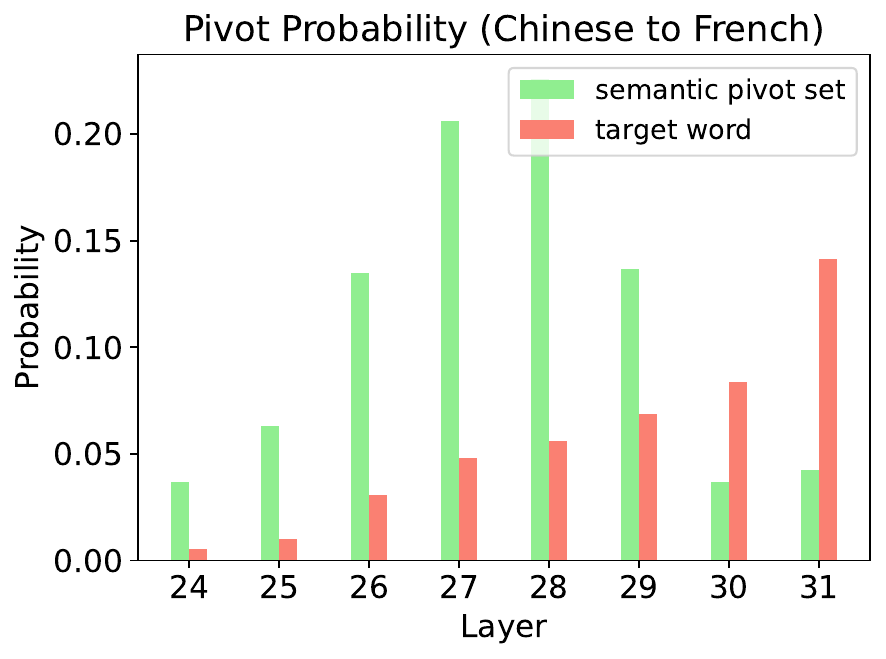}
  \includegraphics[width=0.33\linewidth]{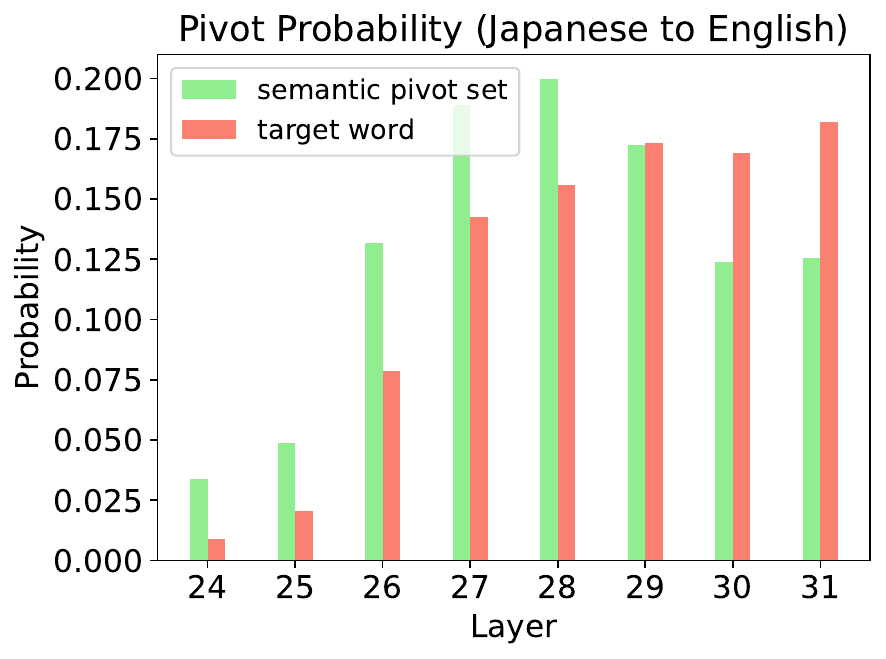}
  \caption{The semantic pivots set of probability in the last eight layers. The x-axis represents the OLMo-7B’s layer index, and the y-axis indicates the total probability of all tokens in the semantic pivots set. The title shows the source language and the target language.}
  \label{fig:pivot}
\end{figure*}
\label{d}
In Section \ref{sec4.3}, we conduct a token co-occurrence proportion analysis, to explore how models learn cross-lingual abilities through semantic pivots. We show the semantic pivot set of probability in the last
eight layers in a part of the language pairs in Figure \ref{fig:pivot}.
\end{document}